\documentclass{article}

\usepackage{subfigure}

\usepackage{spconf}
\usepackage{graphicx}
\usepackage{subfig}
\usepackage[]{geometry}
\usepackage{amsmath}
\usepackage{amsfonts}
\usepackage{listings}
\usepackage{float}
\usepackage{alltt}

\title{Photo-unrealistic Image Enhancement for Subject Placement in Outdoor Photography}
\name{Christian Tendyck\sthanks{Visiting Purdue from Ruhr Universit\"at Bochum, Germany.}, Andrew Haddad,  Mireille Boutin}
\address{School of Electrical and Computer Engineering  \\
	Purdue University\\
	465 Northwestern Ave., West Lafayette, IN 47907 USA}

\begin{document}
\maketitle

\begin{abstract}
Camera display reflections are an issue in bright light situations, as they may prevent users from correctly positioning the subject in the picture. We propose a software solution to this problem, which consists in modifying the image in the viewer, in real time. In our solution, the user is seeing a  posterized image which roughly represents the contour of the objects. Five enhancement methods are compared in a user study. Our results indicate that
the problem considered is a valid one, as users had problems locating landmarks nearly 37\% of the time under sunny conditions, and that our proposed enhancement method using contrasting colors is a practical solution to that problem.
\end{abstract}
%

\section{Introduction}
\begin{figure}[t]
	\centering
		\includegraphics[width=0.5\linewidth]{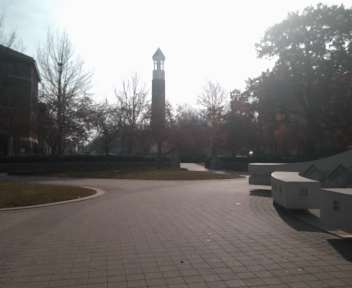}
		\caption{Image Without Enhancement. The user could not see the Bell Tower on the display. }
		\label{fig:Original} 
\end{figure}

Image visibility on a display is influenced by many factors, such as the image contrast, display brightness and physical characteristics of the display hardware. One important issue is ambient/spectral light reflections. 
In some situations, these reflections reduce visibility entirely and make it impossible to see anything on the screen except the reflections. 
The problem of reflections caused by bright sunlight is well known in applications such as automated tellers. People who use a laptop, tablet computer or smart phone outdoors are also familiar with this issue. For camera displays, including smart phone cameras, this can be problematic. Indeed, in order to position the subject in the correct location in the picture, it is necessary to hold the camera with a specific angle and at a specific position. When the display visibility is poor near the desired angle and position, correctly positioning the subject can become nearly impossible (Fig.~\ref{fig:Original}).

As we explain in Section \ref{section:existing}, there are hardware solutions to deal with display reflections.  We are interested in developing an affordable solution for camera phones and other low-cost camera-equipped portable devices running on batteries. Our proposed solution is a software that replaces the images of the camera displayed in the viewer  by ``enhanced"  images, in real time. These enhanced images contain few, highly contrasting colors so to be more easily viewable on the camera display in bright light conditions. The colors of the image represent a coarse segmentation of the scene elements, thus helping the viewer to correctly position the subject in the viewer before taking a picture.   

In the following, we investigate five different low-computation segmentation methods that produce a posterized, non photo-realistic image fast enough so that the images can be displayed in the camera viewer in real-time. These methods are described in Section \ref{section:proposed}. To compare these methods, we performed a user study, described in Section \ref{section:results}. Our results confirm that the problem of locating a subject in a camera viewer is a concern on sunny days.
Indeed, our subjects indicated that they could not locate the pre-specified landmarks on the given camera phone display nearly 37\% of the time 
on sunny days. Cloudy days are less problematic, as the landmark could not be located only once (1 out of 13 cases). When the landmark was not visible without enhancement, at least one of our proposed methods helped the viewer locate the landmark in 77.1\% of cases (27 out of 35 cases). We thus conclude that the problem considered is a valid one, and that our proposed coarse segmentation approach using contrasting colors is a useful, practical, and affordable solution to that problem, though different enhancements may be needed depending on the subject considered and context.

\begin{figure*}[ht]
\centering
\subfigure[Histogram Equalized Image.]{
  \includegraphics[width=0.3\linewidth]{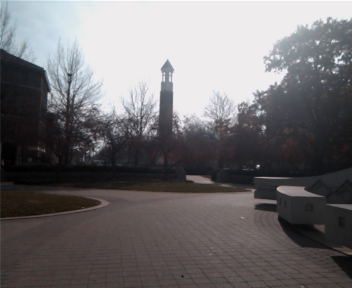}
 }
\subfigure[Gray Scale Thresholding.]{
  \includegraphics[width=0.3\linewidth]{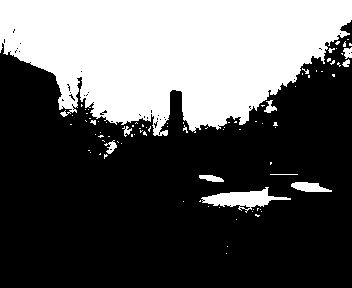}
 }
 \subfigure[Otsu's Gray Scale Thresholding.]{
 \includegraphics[width=0.3\linewidth]{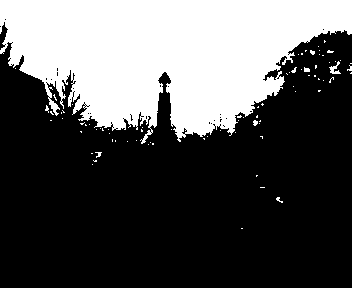}
 }\\
 \subfigure[RGB Thresholding]{
  \includegraphics[width=0.3\linewidth]{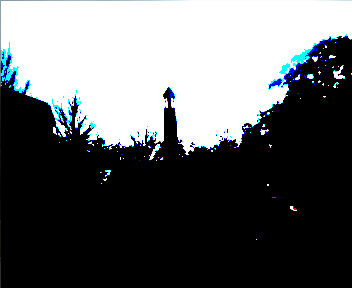}
 }
\subfigure[RGB Max Thresholding.]{
  \includegraphics[width=0.3\linewidth]{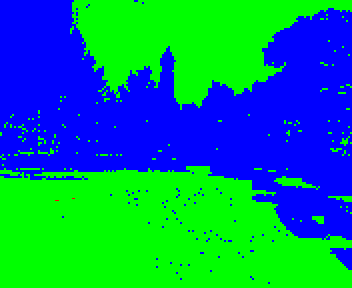}
 }
 \subfigure[Thresholding after Decorrelation.]{
 \includegraphics[width=0.3\linewidth]{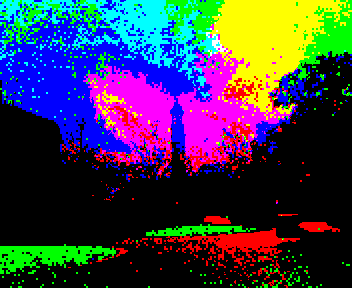}
 }\
\caption{The user still could not see the Bell Tower in a), but could see it in b) c) d) e). While f) was overall our best enhancement method, the user could not see the tower with it in this specific case.   }
\label{fig:methods}
\end{figure*}



\section{Existing methods}
\label{section:existing}

There are hardware solutions to deal with display reflections. 
One is the application of an antireflective coating (Rayleigh's film) to address the reflections due to the difference between the refractive index of the air and the display \cite{Rayleigh}. The idea is to add a coating with a refractive index between the refractive index of the air and that of the glass. 
Another way is to choose the thickness of the coating layer to be $1/4$ of the wavelength that causes the reflections, thus creating destructive interference \cite{Macleod}.
Total destructive interference can be obtained for one wavelength at one angle, but one can eliminate more reflections by adding multiple coatings. 

One popular technique for laptop displays is to add an anti-reflective etch to the screen so to create a ``matte" screen. This is done by treating the display surface with acid. This reduces reflections by scattering light on the screen in all directions equally. However, 
this tend to decrease the image sharpness.

Increasing the brightness of a display increases its contrast, as contrast is proportional to the difference between the luminance of the displayed image and the background luminance.  ``Highbright displays'' do this by emitting light at about 1700-2000 Cd/m$^2$. They adjust  to light conditions using ambient light sensors. But they have high power consumption, a reduced lifetime  backlight performance deteriorates over time \cite{3}.  
Transflective displays use a portion of the ambient light to display images. Thus, they produce brighter illumination with less energy than highbright displays. This reduces running costs and leads to a longer lifetime. Tests have also shown that they achieve better results in environments of bright sunlight compared to highbright displays \cite{3}. 
A more recent technique produces large area coatings with nipple arrays that simulate the surface of a moth eye. The shape and the depth of the nipples can be adjusted during production, thus adjustabling the reflectivity  \cite{4}.



\section{Segmentation Methods investigated}
\label{section:proposed}

Histogram equalization (Fig.~\ref{fig:methods}a) is a standard technique to produces a photo-realistic image with better contrast. To further increase contrast, we propose 5 segmentation methods that yield poster-like images with highly contrasting colors. 

{\bf Method 1: Gray Scale Thresholding}
The first method we will investigate is the simplest: we map the gray scale value of the pixel to either black or white by thresholding in the middle (128). The result is a black/white image (Fig.~\ref{fig:methods}b). A black and white color contrast is likely to be the most robust under light reflections. But using only two colors limits the number of objects that can be separated.

{\bf Method 2: Otsu's Gray Scale Thresholding}
A better way to segment an image is Otsu's method \cite{1}, which determines the threshold by creating a histogram of the gray values in the image and maximizing the separability of the classes. Our second method is based on this relatively low-computation thresholding technique. For simplicity, we set the number of classes to two so to obtain a binary segmentation. This segmentation is then represented, again, by a black and white image. See Fig.~\ref{fig:methods}c.

{\bf Method 3: RGB Thresholding}
Here we threshold the three color channels of the RGB image separately, in the middle (128). As one can see in Fig.~\ref{fig:methods}d, this yields many black or white pixels. This is due to the fact that the color channels are correlated: i is likely that a pixel that has a low/high value in one channel also has low/high values in the other two channels. Thus, many pixels get mapped to either black or white and just a few get mapped to one of the other six possible colors (red, green, blue, yellow, cyan, magenta).


{\bf Method 4: RGB Max Thresholding}
The maximum among the three RGB coordinates of each pixel  is set to 255, and the two remaining coordinates are set to zero. This creates an image with three colors (Red, Green and Blue). See Fig.~\ref{fig:methods}e.

{\bf Method 5: Thresholding after Decorrelation}
 Decorrelation Stretch is often used in remote sensing \cite{2}. The idea is to decorrelate the color channels using a principle component analysis (PCA) transformation. Here we produce a posterized image by thresholding the three color channels separatelly (in the middle) after the PCA transformation. This gives us an eight color images (the corners of the color cube: black, red, green, blue, yellow, cyan, magenta and white). See Fig.~\ref{fig:methods}f.

\begin{table*}
\caption{User Study Results. Rational entries represent the fraction of cases for which users found that the enhancement improved their ability to see the landmark. (Number of corresponding cases are in parenthesis.)}
\begin{center}
\begin{tabular}{|c|c|c|c|c|c|c|}
\hline
Sunny Conditions (93) &&&&&&\\
\hline
Landmark seen  & Hist. Eq. & Thresh. & Otsu & RGB Thresh. & RGB Max  & Thresh.  \\
 w/out enhancement? &  &  &  &  &   & after Decorr.  \\
No (34) & 10/34 & 16/34  & 11/34 & 18/34 & 10/34 & 23/34 \\
\hline
Yes (59) &22/59 & 11/59 &9/59 &15/59 &5/59 & 30/59 \\
\hline
\hline
Cloudy Conditions (13) &&&&&&\\
\hline
Landmark seen  & Hist. Eq. & Thresh. & Otsu & RGB Thresh. & RGB Max  & Thresh.  \\
 w/out enhancement? &  &  &  &  &   & after Decorr.  \\
 \hline
No (1) & 0/1& 0/1 &0/1 & 0/1 &1/1 & 1/1\\
\hline
Yes (12) & 2/12 & 0/12 & 0/12 & 0/12 & 0/12 & 2/12 \\
\hline
\end{tabular}
\end{center}
\label{table:sunnycloudy}
\end{table*}

\section{Experiments}
\label{section:results}
Our user study was performed using an HTC Desire smartphone with a 1GHz Snapdragon Processor, 512 MB ROM, 384 MB RAM, a 3.7 inches WVGA touch screen display with 480 X 800 pixel resolution and a 5 megapixel camera with auto focus. 
We selected nine outdoor landmarks of the Purdue campus for our study (Neil Armstrong statue, Arch, Loeb Playhouse, John Purdue memorial, Purdue Memorial Union, "P" statue, lion head sculpture, Bell Tower, engineering fountain.) Fifteen subjects were recruited. Each subject was brought to the location of some randomly selected landmark and asked to take a picture of the landmarks from a specific location. At first, the subject was shown the original image in the viewer and asked whether they could locate the landmark. Subsequently, the camera displayed an histogram equalized version of the images in the viewer. The subject was then asked whether they could now better see the landmark on the viewer. Then all 5 proposed segmentation methods were used to modify the images in the camera viewer, one after the other, and for each improvement, the user was asked whether the corresponding enhancement was helpful in helping them see the landmark. The sky's lighting conditions were recorded (sunny/cloudy). 

Each user was presented with between 6 and 9 landmarks, for a total of 106 cases (i.e., 106 distinct landmark-user pairs). Of these, 13 cases were under cloudy conditions and the remaining 93 were under sunny conditions.  As shown in Table \ref{table:sunnycloudy}, without enhancement, the user was able to see the landmark specified on the camera viewer in almost all cases when the conditions were cloudy (12 out of 13 cases). In contrast, under sunny conditions, users were unable to see the landmarks in 34 out of 93 cases (about 37\% of the time). So the presence of sun in the sky is indeed an issue. 

When the subject could not locate the landmark initially under sunny condition, there was an improvement from at least one method in 26 out of 34 cases (76.5\%).  Of the 5 proposed enhancements, Threshold after Decorrelation stands out as the best, providing an improvement over the original image in 23 out of these 34 cases (67.6\%). This may be because this method produces a more detailed image. However, there is a tradeoff between details and robustness under reflections. For example, in the case shown in Fig. \ref{fig:methods} f), the user could not see the landmark with this method, while a simple black and white threshold was helpful (Fig.~\ref{fig:methods}).

\section{Conclusion and Future Work}
\label{section:conclusion}
We proposed to help users locate subjects on camera displays in bright light conditions by increasing the contrast of the images using a posterized, non-photorealistic enhancement obtained by segmentation. Five different segmentation methods were proposed for this purpose and we performed a user study to compare them. Our results indicate that, when trying to take pictures under sunny conditions, being able to switch the viewer to any of these 5 enhancement methods could help users locate the subject in the camera viewer.

\bibliographystyle{IEEEbib}
\bibliography{refs}

\end{document}